\documentclass[twocolumn]{article}

\usepackage{arxiv}

\usepackage[utf8]{inputenc} 
\usepackage[T1]{fontenc}    
\usepackage{hyperref}       
\usepackage{url}            
\usepackage{booktabs}       
\usepackage{amsfonts}       
\usepackage{nicefrac}       
\usepackage{microtype}      
\usepackage{lipsum}
\usepackage{graphicx}

\usepackage{natbib}
\usepackage{lastpage}
\usepackage{tikz}
\usepackage[english]{babel}

\usepackage{tkz-graph}
\usepackage{listings}

\usetikzlibrary{
    arrows,
    arrows.meta,
    fadings,
    patterns,
    positioning,
    shapes.geometric
}

\usepackage{xspace}
\usepackage{amsmath}
\usepackage{cleveref}
\usepackage{algorithm}
\usepackage{algpseudocode}
\usepackage{geometry}
\usepackage{siunitx}
\usepackage{bm}
\usepackage{xcolor}
\usepackage{pgfplots}
\usepackage{tabularx}
\usepackage{makecell}
\usepackage{array}

\usepackage[nolist]{acronym}
\begin{acronym}[UML]
    \acro{KG}{knowledge graph}
    \acro{KGE}{Knowledge Graph Embedding}
    \acro{LLM}{Pretrained Large Language Model}
    \acro{RAG}{Retrieval-augmented Generation}
    \acro{OWL}{Web Ontology Language}
    \acro{DLs}{Description Logics}
    \acro{CEL}{Class Expression Learning}
\end{acronym}

\definecolor{tab20darkblue}{HTML}{4e79a7}
\definecolor{tab20darkgreen}{HTML}{59a14f}
\definecolor{tab20darkred}{HTML}{e15759}
\definecolor{tab20darkorange}{HTML}{f28e2b}
\definecolor{tab20darkturquoise}{HTML}{499894}
\definecolor{tab20darkgray}{HTML}{79706e}
\definecolor{tab20darkbrown}{HTML}{9d7660}
\definecolor{tab20darkpurple}{HTML}{b07aa1}
\definecolor{tabl20lighgreen}{HTML}{8cd17d}
\definecolor{tab20lightblue}{HTML}{a0cbd8}

\algtext*{EndFor}
\algtext*{EndIf}
\algtext*{EndWhile}
\algtext*{EndFunction}
\pgfplotsset{compat=1.18} 

\usetikzlibrary{arrows,topaths,calc}
\usetikzlibrary{positioning}
\usetikzlibrary{shadows}
\usetikzlibrary{shapes}
\usetikzlibrary{backgrounds}
\usetikzlibrary{decorations.pathreplacing}

\graphicspath{ {./images/} }

\title{Syntactic Simplification of OWL Class Expressions}

\author{
 Alkid Baci, \ N'Dah Jean Kouagou, \ Caglar Demir, \ Axel-Cyrille Ngonga Ngomo \\
  Department of Computer Science \\
  Paderborn University \\
  Warburger Str. 100, 33098 Paderborn, Germany \\
  \texttt{\{alkid.baci, ndah.jean.kouagou, caglar.demir, axel.ngonga\}@upb.de} \\
}

\begin{document}

\twocolumn[
\maketitle
\begin{abstract}
\ac{CEL} often produces complex OWL class expressions that are difficult to interpret and reason over. However, by following theoretically grounded simplification principles, this complexity can be reduced. In this paper, we propose \textbf{Class Expression Simplifier (CES)}, a novel algorithm for the syntactic simplification of class expressions in \ac{DLs}. CES aims to preserve formal semantics while reducing representational complexity. It systematically applies rewriting rules to eliminate redundancies and identify simpler yet equivalent expressions, thereby producing more compact and human-readable representations without altering logical entailments. We evaluate the effectiveness of CES on class expressions learned from two medium-sized ontologies, demonstrating measurable improvements in reasoning efficiency and reductions in verbosity. This work contributes to the broader goal of making ontology-driven applications more accessible, maintainable, and scalable, with direct implications for knowledge graph construction, semantic search, and Web-scale reasoning. CES is implemented within the open-source Python framework \texttt{OWLAPY} \cite{baci2025owlapypythonicframeworkowl} and is publicly available.
\end{abstract}

\vspace{1em}
]

\section{Introduction}
Ontologies expressed in the 
\ac{OWL} \cite{antoniou2009web} have become a cornerstone for representing structured knowledge on the Semantic Web \cite{bernerslee2001semantic}, powering applications ranging from biomedical data integration to knowledge graph construction and Web-scale reasoning \cite{lamy2017owlready, demir2025ontolearn}. At the heart of \ac{OWL} are class expressions, which allow the specification of rich and precise concepts by combining logical operators and role restrictions. While this expressiveness (e.g. $\mathcal{SROIQ^{(D)}}$) is essential for capturing domain semantics, it often leads to complex and verbose expressions \cite{pileggi2019analysing}. 
Such expressions are not only difficult for ontology engineers to interpret and maintain, but can also introduce inefficiencies in automated reasoning and query answering tasks \cite{pileggi2019analysing, kang2012rigorous, lehmann2011class}.

The problem of complex class expressions manifests in several ways. From a usability perspective, verbose or redundant expressions reduce the readability of ontologies \cite{blumer1987occam}, making collaborative development, reuse and verbalization (e.g. via LLMs) challenging. From a computational perspective, the presence of unnecessary syntactic constructs can increase the size of reasoning tasks, negatively affecting the performance of \ac{DLs} reasoners \cite{kang2012rigorous, kang2012predicting}. Simplifying class expressions without altering their semantics is therefore a desirable step to improve both the human- and machine-oriented aspects of ontology use.
In Web-scale applications, where ontologies are combined with massive knowledge graphs, even small inefficiencies in reasoning can become prohibitive \cite{zghal2026co}.

Existing work has primarily focused on semantic reasoning \cite{abburu2012survey}, optimization techniques \cite{abburu2012survey, letelier2013static}, or ontology modularization \cite{leclair2022review, rector2003modularisation}. By contrast, systematic methods for the syntactic simplification of \ac{OWL} class expressions remain largely underexplored. While FaCT++ \cite{tsarkov2006fact} provides basic simplification like for example, transforming expressions into a simplified normal form (SNF), it represents the closest-yet still limited-point of comparison to our approach.

\textbf{Contributions.} In this paper, we address the challenge of simplifying \ac{OWL} class expressions by introducing a syntactic rewriting approach rooted in $\mathcal{SOIQ^{(D)}}$ (restricted to transitive roles and excluding all other RBox axioms), as we adhere strictly to syntactic simplification.
Specifically, we:
\begin{itemize}
    \item [\textbf{(C1)}] \textit{Define a set of DLs-based rewriting rules for syntactic simplification over concepts in $\mathcal{SOIQ^{(D)}}$ with restriction to transitive roles.}
    \item [\textbf{(C2)}] \textit{Propose CES, a rewriting-based algorithm for syntactic simplification of class expressions, following a recursive approach that guarantees termination.}
    \item [\textbf{(C3)}] \textit{Display the effectiveness of CES in reduction of length and reasoning runtime on 200 class expressions generated for two medium-sized datasets.}
\end{itemize}

\section{Background}
OWL ontologies are grounded in \ac{DLs}, where complex class expressions are constructed using boolean connectives, quantifiers, and role restrictions.
Throughout this paper we use the terms `concept' and `class expression' interchangeably because \ac{DLs} concepts can be represented in \ac{OWL} class expressions and vice-versa. In this section, we define the simplification problem and present the simplification principles.

\subsection{Description Logics}

\ac{DLs} provide the formal foundation of \ac{OWL} and define concepts using constructors such as conjunction ($\sqcap$), disjunction ($\sqcup$), negation ($\neg$), existential ($\exists r.C$), and universal ($\forall r.C$) restrictions over roles $r$. These constructs enable the definition of complex concepts that describe sets of individuals satisfying specific conditions. Formally, each DL expression is interpreted under a model $\mathcal{I} = (\Delta^{\mathcal{I}}, \cdot^{\mathcal{I}})$, where $\Delta^{\mathcal{I}}$ is the domain and $\cdot^{\mathcal{I}}$ an interpretation function mapping concepts and roles to subsets of $\Delta^{\mathcal{I}}$ and binary relations, respectively. 

\subsection{Principles of Syntactic Simplification}
Towards the goal of syntactic simplification we propose the following principles:
\begin{itemize}
    \item \textbf{Redundancy Elimination:} remove duplicates of subexpressions or order/flatten conjuncts and disjuncts.
    \item \textbf{Equivalence Preservation:} apply algebraic laws/rules to produce shorter but semantically identical expressions.
\end{itemize}
These principles guide the design of our rewriting algorithm, ensuring termination and semantic soundness. Below, we describe some of the most profound rules applied to simplify class expressions under the equivalence preservation principle. Our function processes expressions according to the \textit{unique name assumption} (UNA).
\paragraph{Absorption} 
The absorption law eliminates redundant disjunctive or conjunctive constructs by exploiting logical inclusion. Formally, for any class expressions $C$ and $D$, the following equivalences hold:

\[
C \sqcup (C \sqcap D) \equiv C \quad \text{and} \quad C \sqcap (C \sqcup D) \equiv C.
\]

\paragraph{Idempotence} 
The idempotence law ensures that repeated occurrences of the same class expression within a conjunction or disjunction do not increase its meaning:
\[
C \sqcap C \equiv C \quad \text{and} \quad C \sqcup C \equiv C.
\]
\paragraph{Identity and Domination} 
The identity and domination laws describe the interaction of class expressions with the universal concept $\top$ and the empty concept $\bot$. The identity law captures neutral behavior while the domination law expresses overriding behavior:
$$
 C \sqcap \top \equiv C, \quad C \sqcup \bot \equiv C, \quad 
 C \sqcup \top \equiv \top, \quad C \sqcap \bot \equiv \bot .
$$
\paragraph{Law of the Excluded Middle and Law of Non-Contradiction} 
The law of the excluded middle expresses that, for any class expression $C$, either $C$ or its negation must hold universally whereas the law of non-contradiction states that a class expression and its negation cannot be simultaneously satisfied. Respectively:
$$
C \sqcup \neg C \equiv \top \quad \text{and} \quad C \sqcap \neg C \equiv \bot.
$$

\paragraph{Quantifier Distribution} 
Quantifier distribution combines multiple universal or existential restrictions over the same role ($r$) into a single restriction with a merged filler expression:
$$
\forall r.C \sqcap \forall r.D \equiv \forall r.(C \sqcap D)
\quad \text{and} $$
$$
\exists r.C \sqcup \exists r.D \equiv \exists r.(C \sqcup D).
$$
\paragraph{Cardinality Restriction Subsumption} 
When multiple cardinality restrictions share the same role and filler, redundant constraints can be eliminated by selecting the less or more restrictive bound, depending on the connective. Formally, for a role $r$, a class expression $C$, and integers $m,n$ with $m < n$:
$$(\geq m\, r.C) \sqcup (\geq n\, r.C) \equiv (\geq m\, r.C)
\quad \text{and}$$
$$
(\geq m\, r.C) \sqcap (\geq n\, r.C) \equiv (\geq n\, r.C).
$$
Analogous equivalences are used for upper-bound restrictions ($\le$) as well as for datatype restrictions.
\paragraph{Factorization} 
Factorization exploits distributivity to extract common subexpressions from disjunctions or conjunctions, reducing repetition and structural complexity:
$$
(C \sqcap D) \sqcup (C \sqcap E) \equiv C \sqcap (D \sqcup E)
\quad \text{and}
$$
$$
(C \sqcup D) \sqcap (C \sqcup E) \equiv C \sqcup (D \sqcap E).
$$
Factorization is also applied to negated concepts to factor out the negation in order to reduce the amount of negation constructs used to represent the complex concept.

\section{Methodology}

We define \textit{syntactic simplification} as a transformation function $f$ on the class expression $\mathcal{C}$ such that $f(C)$ is syntactically simpler than $C$ in terms of length--i.e. \ac{OWL} constructs required to represent the expression \cite{lehmann2011class}, and semantically equivalent, i.e., $C^{\mathcal{I}} = f(C)^{\mathcal{I}}$ for all interpretations $\mathcal{I}$.

When handling complex class expressions, we often encounter combinations of n-ary boolean connectives (i.e., conjunctions and disjunctions of concepts). For brevity, we refer to these simply as n-ary expressions and to the concepts in such expressions as \textit{operands}. Our simplification algorithm is particularly effective in this setting, since a larger number of simplification rules can typically be applied. In contrast, when expressions lack such connectives, the simplification process is naturally less extensive.
 
The objective of the algorithm is to apply as many rules within a single simplification pass  as possible (i.e., one execution of the algorithm on a given class expression). To this end, we implemented a recursive function $\texttt{simplify}(c, p)$, where $c$ denotes the current class expression and $p$ denotes its parent n-ary expression. The second parameter is essential for rules that require knowledge of the surrounding context, i.e., the class expression that is one level higher in the recursion. The \texttt{simplify} function is implemented as a single-dispatch function, with different specializations for each type of class expression, which makes recursive traversal both natural and efficient.

\subsection{Simplification of N-ary Boolean Expressions}

For n-ary expressions, simplification proceeds both at the operand level and at the connective level. In Algorithm \ref{alg:simplify_union}, we show how the \texttt{simplify} function is implemented for 
\textit{ObjectUnionOf}. Nested n-ary expressions are first flattened (e.g., $(C \sqcap (D \sqcap E)) \to (C \sqcap D \sqcap E)$) to facilitate rule application, and then the \textit{Law of Idempotence }is applied to remove repeated operands. During recursion, \textit{Absorption Law} is applied when common elements exist between an expression and its parent, and duplicates are removed using set-based storage. In case the $\top$ or $\bot$ concepts occur in the set of operands we apply the \textit{Law of Domination/Identity}. The algorithm also applies the \textit{Law of the Excluded Middle}, removing operands that appear alongside their negations, and handles  \textit{subsumption checks} for cardinality restrictions. Finally, \textit{factorization} is applied via \texttt{apply\_factorization} to reduce structural length. 

The simplify function for intersection (\texttt{ObjectIntersectionOf}) follows a similar procedure, with slight differences corresponding to the specific rules applicable to conjunctions. Some minor implementation details have been omitted in Algorithm \ref{alg:simplify_union} for clarity, but the overall structure ensures a systematic, recursive application of all relevant simplification rules in a single pass.

\begin{algorithm}
\caption{Recursive function $\mathsf{simplify}(c, p)$ for \texttt{ObjectUnionOf}} 
\label{alg:simplify_union}
\begin{algorithmic}[1]
\State $c' \gets \mathsf{combine\_nary\_expressions}(c)$ \Comment{Flatten nested unions \& apply Idempotence law}
\If{$c' \neq c$} 
    \State \Return $\mathsf{simplify}(c', p)$
\EndIf

\State $O_c \gets \mathsf{operands}(c)$
\If{$p \neq \text{Null}$} 
    \State Apply Absorption law: remove common operands with parent $p$ (if any)
\EndIf

\State $S \gets \mathsf{map}(\mathsf{simplify}, O_c, \mathsf{repeat}(c))$ \Comment{Recursive simplification of operands}

\If{$|S| = 1$} \State \Return $\mathsf{pop}(S)$ \EndIf
\If{$\top \in S$} \State \Return $\top$ \Comment{Domination law} \EndIf
\If{$\bot \in S$} \State $S \gets S \setminus \{\bot\}$ \Comment{Identity law} \EndIf

\For{$e \in \mathsf{copy}(S)$} 
    \If{$\mathsf{isinstance}(e, \text{ObjectComplementOf})$ and operand $o_e \in S$}
        \State Remove $\{e, o_e\}$ \Comment{Law of the excluded middle}
        \If{$S = \emptyset$} \State \Return $\top$ \EndIf
        \If{$|S| = 1$} \State \Return $\mathsf{pop}(S)$ \EndIf
    \EndIf
\EndFor

\State $c \gets \text{new ObjectUnionOf}(S)$
\For{$o \in \mathsf{operands}(c)$ if $o$ is a cardinality restriction}
    \State Merge cardinality restrictions on the same role and filler
    \State Update $c \gets \text{new ObjectUnionOf}(S)$
    \State \textbf{break}
\EndFor

\If{$p = \text{Null}$} \Comment{Root expression}
    \State $c \gets \mathsf{apply\_factorization}(c, \text{True})$
\EndIf

\State \Return $c$
\end{algorithmic}
\end{algorithm}

\subsection{Factorization}
\label{sec:factorization}

Factorization is important when it comes to reducing length of the expressions. 
Throughout the simplification process, intermediate class expressions may remain structurally verbose, even after the application of local rewriting rules and can further benefit from factorization which we apply at the root level of recursive union or intersection simplifications, following the simplification of all operands. We consider this an important step since factorization can substantially reduce the number of OWL constructs required to represent a concept. The high-level pseudocode of the factorization procedure is presented in Algorithm~\ref{alg:factorization}.

Factorization is implemented recursively via the recursive function $\texttt{apply\_factorization}(c, \text{first\_iteration})$. On the first call from the simplify function, the class expression $c$ is converted into its top-level \textit{disjunctive normal form} (DNF). This ensures that all disjunctions are explicitly represented at the outermost level, facilitating the identification of redundant or common subexpressions and reducing repeated traversal of nested structures. 
Furthermore, in DNF, redundant disjuncts are easier to identify because they appear as explicit top-level components rather than being hidden inside nested structures. 
To handle both unions and intersections uniformly, two dynamic variables, \texttt{type\_a} and \texttt{type\_b} are assignd the types ObjectUnionOf and ObjectIntersectionOf where \texttt{type\_a} is equal to the type of the class expression $c$ and \texttt{type\_b} is the remaining one. Operands of $c$ are partitioned into those of \texttt{type\_b} $O_{c,\text{type\_b}}$,
and the rest $O_{c, \overline{\text{type\_b}}}$. The \textit{Absorption Law }is applied when both sets are non-empty. For multiple operands of type \texttt{type\_b}, common components are identified pairwise, and local factorization is performed, which is then merged with the remaining operands. Depending on the sizes of the operand sets, the function returns either the locally factorized expression or the combination of factorized and remaining operands. Single operands of type \texttt{type\_b} are recursively factorized before being combined with the rest.

This recursive design ensures that factorization proceeds from general to specific cases, repeatedly applying simplification until no further factorization is possible. The final result is a structurally compact class expression that preserves the semantics of the original expression.

\begin{algorithm}
\caption{Function $\mathsf{apply\_factorization}(c, \text{first\_iteration})$ for n-ary expressions}
\label{alg:factorization}
\begin{algorithmic}[1]
\If{$\text{first\_iteration}$}
    \State $c \gets \mathsf{get\_top\_level\_dnf}(c)$
\EndIf
\State Determine \texttt{type\_a} and \texttt{type\_b} based on $c$
\State $O_c \gets \mathsf{operands}(c)$
\State $O_{c,\text{type\_b}} \gets \{ o \in O_c \mid \mathsf{isinstance}(o, \text{type\_b}) \}$
\State $O_{c, \overline{\text{type\_b}}} \gets O_c \setminus O_{c,\text{type\_b}}$
\If{$O_{c,\text{type\_b}} \neq \emptyset$ and $O_{c, \overline{\text{type\_b}}} \neq \emptyset$}
    \State Apply Absorption between sets
\EndIf
\If{$|O_{c,\text{type\_b}}| \ge 2$}
    \For{each pair in $O_{c,\text{type\_b}}$}
        \State Identify common operands and perform local factorization
        \State Merge factorized results with remaining operands
    \EndFor
\ElsIf{$|O_{c,\text{type\_b}}| = 1$}
    \State Recursively factorize the single operand and combine with $O_{c, \overline{\text{type\_b}}}$
\EndIf
\State \Return $c$
\end{algorithmic}
\end{algorithm}

\section{Evaluation}

\begin{figure}[tb]
    \centering
    \tikzset{font=\small\fontfamily{phv}\selectfont}
    \begin{tikzpicture}[scale=0.77]
        \begin{axis}[
            title={Mutagenesis},
            width=5.5cm,
			height=5cm,
            ylabel={Length (construction units)},
            grid=both,
            legend style={
                draw=none,
                at={(0,1)},
                anchor=north west,
                font=\small
            },
            xmin=0, xmax=100,
            ymin=0, ymax=4500,
            ytick={ 0, 500, 1500, 2500, 3500, 4500},
            tick style={black},
        ]

        \addplot[
            thick,
            blue,
            mark=*,
            mark size=0.8pt,
        ]
        coordinates {
            (1, 2) (2, 2) (3, 4) (4, 4) (5, 12) (6, 12) (7, 13) (8, 13) (9, 16) (10, 17) (11, 22) (12, 26) (13, 28) (14, 33) (15, 40) (16, 46) (17, 50) (18, 55) (19, 58) (20, 58) (21, 58) (22, 70) (23, 83) (24, 84) (25, 87) (26, 96) (27, 96) (28, 112) (29, 131) (30, 137) (31, 138) (32, 147) (33, 153) (34, 166) (35, 185) (36, 211) (37, 211) (38, 211) (39, 216) (40, 247) (41, 254) (42, 287) (43, 296) (44, 302) (45, 305) (46, 306) (47, 313) (48, 326) (49, 333) (50, 334) (51, 375) (52, 376) (53, 380) (54, 381) (55, 406) (56, 413) (57, 441) (58, 458) (59, 495) (60, 548) (61, 597) (62, 622) (63, 644) (64, 648) (65, 679) (66, 679) (67, 680) (68, 698) (69, 729) (70, 741) (71, 833) (72, 872) (73, 885) (74, 898) (75, 913) (76, 939) (77, 953) (78, 966) (79, 997) (80, 1003) (81, 1008) (82, 1033) (83, 1088) (84, 1188) (85, 1236) (86, 1265) (87, 1510) (88, 1532) (89, 1581) (90, 1696) (91, 1819) (92, 1840) (93, 1941) (94, 1952) (95, 2051) (96, 2100) (97, 2161) (98, 2467) (99, 2782) (100, 4148)
        };
        \addlegendentry{Original}

        \addplot[
            thick,
            red,
            mark=square*,
            mark size=0.8pt,
        ]
        coordinates {
            (1, 2) (2, 2) (3, 4) (4, 4) (5, 12) (6, 12) (7, 13) (8, 13) (9, 13) (10, 17) (11, 22) (12, 21) (13, 27) (14, 22) (15, 38) (16, 46) (17, 46) (18, 55) (19, 39) (20, 45) (21, 40) (22, 37) (23, 57) (24, 44) (25, 32) (26, 59) (27, 51) (28, 67) (29, 82) (30, 71) (31, 117) (32, 53) (33, 95) (34, 87) (35, 81) (36, 65) (37, 91) (38, 100) (39, 108) (40, 135) (41, 93) (42, 69) (43, 122) (44, 118) (45, 271) (46, 176) (47, 206) (48, 196) (49, 162) (50, 119) (51, 281) (52, 132) (53, 188) (54, 116) (55, 193) (56, 122) (57, 238) (58, 300) (59, 261) (60, 307) (61, 338) (62, 278) (63, 305) (64, 185) (65, 479) (66, 172) (67, 227) (68, 382) (69, 363) (70, 335) (71, 202) (72, 486) (73, 367) (74, 284) (75, 577) (76, 468) (77, 299) (78, 504) (79, 614) (80, 464) (81, 433) (82, 366) (83, 428) (84, 622) (85, 570) (86, 634) (87, 1009) (88, 787) (89, 781) (90, 1001) (91, 597) (92, 1115) (93, 453) (94, 949) (95, 1136) (96, 280) (97, 661) (98, 1256) (99, 1501) (100, 1983)
        };
        \addlegendentry{CES}

        \end{axis}
    \end{tikzpicture}
    \begin{tikzpicture}[scale=0.77]
        \begin{axis}[
            title={Carcinogenesis},
            width=5.5cm,
			height=5cm,
            grid=both,
            legend style={draw=none, at={(0,1)}, anchor=north west,       font=\small},
            xmin=0, xmax=100,
            ymin=0, ymax=4500,
            ytick={ 0, 500, 1500, 2500, 3500, 4500},
            tick style={black},
            yticklabels={}, 
        ]

        \addplot[
            thick,
            blue,
            mark=*,
            mark size=0.8pt,
        ]
        coordinates {
            (1, 2) (2, 2) (3, 4) (4, 10) (5, 12) (6, 15) (7, 21) (8, 23) (9, 24) (10, 27) (11, 37) (12, 39) (13, 47) (14, 52) (15, 57) (16, 57) (17, 59) (18, 60) (19, 61) (20, 64) (21, 68) (22, 70) (23, 77) (24, 84) (25, 86) (26, 99) (27, 102) (28, 117) (29, 118) (30, 138) (31, 157) (32, 173) (33, 186) (34, 188) (35, 192) (36, 209) (37, 223) (38, 229) (39, 230) (40, 242) (41, 245) (42, 255) (43, 259) (44, 260) (45, 274) (46, 275) (47, 282) (48, 291) (49, 336) (50, 337) (51, 347) (52, 427) (53, 440) (54, 447) (55, 465) (56, 470) (57, 486) (58, 513) (59, 516) (60, 544) (61, 553) (62, 565) (63, 571) (64, 595) (65, 646) (66, 655) (67, 676) (68, 713) (69, 722) (70, 731) (71, 747) (72, 757) (73, 782) (74, 802) (75, 831) (76, 930) (77, 951) (78, 974) (79, 977) (80, 1131) (81, 1147) (82, 1243) (83, 1449) (84, 1483) (85, 1527) (86, 1607) (87, 1620) (88, 1891) (89, 1956) (90, 2029) (91, 2055) (92, 2223) (93, 2247) (94, 2338) (95, 2385) (96, 2800) (97, 3272) (98, 3399) (99, 3439) (100, 4284)
        };
        \addlegendentry{Original}

        \addplot[
            thick,
            red,
            mark=square*,
            mark size=0.8pt,
        ]
        coordinates {
            (1, 2) (2, 2) (3, 4) (4, 10) (5, 12) (6, 15) (7, 18) (8, 22) (9, 19) (10, 24) (11, 36) (12, 24) (13, 36) (14, 39) (15, 38) (16, 51) (17, 33) (18, 36) (19, 31) (20, 55) (21, 64) (22, 46) (23, 50) (24, 59) (25, 85) (26, 91) (27, 81) (28, 44) (29, 53) (30, 72) (31, 84) (32, 78) (33, 90) (34, 109) (35, 77) (36, 91) (37, 117) (38, 86) (39, 106) (40, 146) (41, 118) (42, 97) (43, 156) (44, 140) (45, 160) (46, 104) (47, 65) (48, 148) (49, 117) (50, 112) (51, 227) (52, 274) (53, 253) (54, 256) (55, 339) (56, 114) (57, 227) (58, 130) (59, 110) (60, 353) (61, 293) (62, 312) (63, 183) (64, 313) (65, 246) (66, 239) (67, 501) (68, 235) (69, 294) (70, 472) (71, 144) (72, 326) (73, 430) (74, 388) (75, 454) (76, 459) (77, 484) (78, 559) (79, 549) (80, 349) (81, 697) (82, 464) (83, 526) (84, 539) (85, 694) (86, 626) (87, 623) (88, 1031) (89, 708) (90, 516) (91, 1247) (92, 610) (93, 1072) (94, 1250) (95, 1329) (96, 873) (97, 1198) (98, 1621) (99, 1030) (100, 2485)
        };
        \addlegendentry{CES}

        \end{axis}
    \end{tikzpicture}
    \begin{tikzpicture}[scale=0.77]
        \begin{axis}[
            width=5.5cm,
			height=5cm,
            xlabel={CE index},
            grid=both,
            legend style={draw=none, at={(0,1)}, anchor=north west,       font=\small},
            ylabel={Instance retrieval runtime (s)},
            xmin=0, xmax=100,
            ymin=0, ymax=105,
            ytick={ 0, 20, 40, 60, 80, 100},
            tick style={black},
        ]

        \addplot[
            thick,
            gray,
            mark=*,
            mark size=0.8pt,
        ]
        coordinates {
            (1, 0.05743288993835449) (2, 0.05762481689453125) (3, 0.058435916900634766) (4, 0.05865812301635742) (5, 0.06082010269165039) (6, 0.06692790985107422) (7, 0.11438655853271484) (8, 0.1650388240814209) (9, 0.22794270515441895) (10, 0.2749788761138916) (11, 0.318941593170166) (12, 0.3278641700744629) (13, 0.3371245861053467) (14, 0.4064900875091553) (15, 0.42252302169799805) (16, 0.5029685497283936) (17, 0.5392670631408691) (18, 0.6358709335327148) (19, 0.662822961807251) (20, 0.6697266101837158) (21, 0.8632655143737793) (22, 0.9755635261535645) (23, 1.130265474319458) (24, 1.1347908973693848) (25, 1.141047716140747) (26, 1.1762633323669434) (27, 1.3032095432281494) (28, 1.518287181854248) (29, 1.6701719760894775) (30, 1.6929256916046143) (31, 1.8268115520477295) (32, 1.852782964706421) (33, 2.0429210662841797) (34, 2.0765674114227295) (35, 2.322726249694824) (36, 2.4500670433044434) (37, 2.6139490604400635) (38, 2.8442540168762207) (39, 2.945335626602173) (40, 3.0661697387695312) (41, 3.352247714996338) (42, 3.3604936599731445) (43, 3.856464385986328) (44, 3.8808562755584717) (45, 3.8965330123901367) (46, 3.946289539337158) (47, 4.002954721450806) (48, 4.069582939147949) (49, 4.174375534057617) (50, 4.49939489364624) (51, 4.500753402709961) (52, 4.886188745498657) (53, 4.913910865783691) (54, 5.1026999950408936) (55, 5.229230642318726) (56, 5.476718187332153) (57, 5.5138373374938965) (58, 5.803210258483887) (59, 6.371552467346191) (60, 7.189193964004517) (61, 7.73660135269165) (62, 7.945321798324585) (63, 8.001362800598145) (64, 8.178551435470581) (65, 8.58963918685913) (66, 8.599650382995605) (67, 8.741483926773071) (68, 9.007812976837158) (69, 9.016621589660645) (70, 10.230152130126953) (71, 10.894164323806763) (72, 10.998024225234985) (73, 11.022801160812378) (74, 11.501392126083374) (75, 11.54869818687439) (76, 12.118462324142456) (77, 12.182649612426758) (78, 12.404802799224854) (79, 12.704142332077026) (80, 12.865407466888428) (81, 13.53258752822876) (82, 13.856517553329468) (83, 14.32052206993103) (84, 15.238126754760742) (85, 15.559008598327637) (86, 17.64691138267517) (87, 18.57546854019165) (88, 18.963738918304443) (89, 20.270793437957764) (90, 22.52218770980835) (91, 22.656337022781372) (92, 23.67639183998108) (93, 25.916239261627197) (94, 26.103612422943115) (95, 27.06462264060974) (96, 27.50778341293335) (97, 29.225589752197266) (98, 33.382160902023315) (99, 34.464508295059204) (100, 51.696226596832275)
        };
        \addlegendentry{Original}

        \addplot[
            thick,
            orange,
            mark=square*,
            mark size=0.8pt,
        ]
        coordinates {
            (1, 0.05858612060546875) (2, 0.05716252326965332) (3, 0.06285810470581055) (4, 0.20349979400634766) (5, 0.059961557388305664) (6, 0.07460570335388184) (7, 0.11141324043273926) (8, 0.10897183418273926) (9, 0.22589635848999023) (10, 0.27409791946411133) (11, 0.11346626281738281) (12, 0.1812887191772461) (13, 0.3345205783843994) (14, 0.5710885524749756) (15, 0.41982269287109375) (16, 0.15865254402160645) (17, 0.23732233047485352) (18, 0.49428272247314453) (19, 0.5935080051422119) (20, 0.8483743667602539) (21, 0.44882941246032715) (22, 0.4451022148132324) (23, 0.6467151641845703) (24, 0.45411181449890137) (25, 0.7100872993469238) (26, 0.493497371673584) (27, 0.5338325500488281) (28, 0.574307918548584) (29, 0.7410283088684082) (30, 0.9921417236328125) (31, 0.41563987731933594) (32, 1.673083782196045) (33, 0.8158891201019287) (34, 0.6180098056793213) (35, 1.1931943893432617) (36, 0.9952545166015625) (37, 0.99114990234375) (38, 1.4416699409484863) (39, 1.4262802600860596) (40, 0.7842972278594971) (41, 0.993401050567627) (42, 0.8511307239532471) (43, 3.520946979522705) (44, 2.261678695678711) (45, 0.6467163562774658) (46, 1.3278217315673828) (47, 2.0718250274658203) (48, 0.9304230213165283) (49, 1.913835048675537) (50, 2.7563071250915527) (51, 0.7846972942352295) (52, 3.5374608039855957) (53, 0.9249422550201416) (54, 1.8641836643218994) (55, 2.489919424057007) (56, 3.530339479446411) (57, 2.229177713394165) (58, 2.9313549995422363) (59, 3.00046706199646) (60, 3.176739454269409) (61, 1.6780693531036377) (62, 3.8678297996520996) (63, 1.2281451225280762) (64, 4.476530075073242) (65, 3.753483533859253) (66, 4.248302936553955) (67, 2.1393938064575195) (68, 6.236496210098267) (69, 4.326355695724487) (70, 4.0949132442474365) (71, 6.124311208724976) (72, 1.9395384788513184) (73, 1.704160451889038) (74, 7.344038724899292) (75, 4.411048889160156) (76, 5.998047828674316) (77, 2.721329927444458) (78, 6.263786554336548) (79, 8.210358381271362) (80, 2.5555083751678467) (81, 6.018833637237549) (82, 5.883307933807373) (83, 3.3970189094543457) (84, 6.48939323425293) (85, 7.517645835876465) (86, 8.390595436096191) (87, 12.676086664199829) (88, 9.728078603744507) (89, 10.301548719406128) (90, 4.36941123008728) (91, 13.125251531600952) (92, 13.971574544906616) (93, 5.176518201828003) (94, 12.130750179290771) (95, 14.24282717704773) (96, 2.652606248855591) (97, 4.975062608718872) (98, 17.256652116775513) (99, 18.367511987686157) (100, 24.610403537750244)
        };
        \addlegendentry{CES}

        \end{axis}
    \end{tikzpicture}
    \begin{tikzpicture}[scale=0.77]
        \begin{axis}[
            width=5.5cm,
			height=5cm,
            xlabel={CE index},
            grid=both,
            legend style={draw=none, at={(0,1)}, anchor=north west,       font=\small},
            xmin=0, xmax=100,
            ymin=0, ymax=105,
            ytick={ 0, 20, 40, 60, 80, 100},
            tick style={black},
            yticklabels={}, 
        ]
        \addplot[
            thick,
            gray,
            mark=*,
            mark size=0.8pt,
        ]
        coordinates {
            (1, 0.0924685001373291) (2, 0.09667515754699707) (3, 0.24146437644958496) (4, 0.2459430694580078) (5, 0.3371396064758301) (6, 0.3712460994720459) (7, 0.44684314727783203) (8, 0.45146822929382324) (9, 0.4892387390136719) (10, 0.5564262866973877) (11, 0.5661461353302002) (12, 0.6327531337738037) (13, 0.8250522613525391) (14, 0.90297532081604) (15, 1.1072704792022705) (16, 1.1234862804412842) (17, 1.1295437812805176) (18, 1.162172555923462) (19, 1.2569293975830078) (20, 1.3012394905090332) (21, 1.3368051052093506) (22, 1.3760952949523926) (23, 1.4717590808868408) (24, 1.5284969806671143) (25, 1.7701056003570557) (26, 1.8014187812805176) (27, 2.1130242347717285) (28, 2.360736608505249) (29, 2.472064733505249) (30, 3.075671911239624) (31, 3.150824785232544) (32, 3.8900136947631836) (33, 4.018527984619141) (34, 4.40228009223938) (35, 4.499768495559692) (36, 4.621992349624634) (37, 4.648873805999756) (38, 4.718406677246094) (39, 4.7354819774627686) (40, 5.167532682418823) (41, 5.176167964935303) (42, 5.625898599624634) (43, 5.639538288116455) (44, 5.692160367965698) (45, 5.749509811401367) (46, 5.873877763748169) (47, 6.116274356842041) (48, 7.192327976226807) (49, 7.635260105133057) (50, 7.995722770690918) (51, 9.044074058532715) (52, 9.17158031463623) (53, 9.91681981086731) (54, 9.917607307434082) (55, 10.134526491165161) (56, 10.574150323867798) (57, 11.01969313621521) (58, 11.633406162261963) (59, 11.864641189575195) (60, 12.012792587280273) (61, 12.044687509536743) (62, 12.132323265075684) (63, 12.341109275817871) (64, 12.708096981048584) (65, 14.001132488250732) (66, 14.487054347991943) (67, 15.675984382629395) (68, 15.69668197631836) (69, 15.976415157318115) (70, 15.996913194656372) (71, 16.490620851516724) (72, 16.654877185821533) (73, 17.33455228805542) (74, 17.958589553833008) (75, 18.41979956626892) (76, 21.32929515838623) (77, 21.83205008506775) (78, 22.77968144416809) (79, 23.028532028198242) (80, 24.87354588508606) (81, 27.015901803970337) (82, 28.51478409767151) (83, 31.576822519302368) (84, 32.820265769958496) (85, 35.61409616470337) (86, 36.529091596603394) (87, 39.01306676864624) (88, 42.404824018478394) (89, 43.181671142578125) (90, 45.60484290122986) (91, 50.2043342590332) (92, 50.2979850769043) (93, 50.48492074012756) (94, 51.72653841972351) (95, 53.00099325180054) (96, 64.49070286750793) (97, 71.30873990058899) (98, 79.7809100151062) (99, 80.81931638717651) (100, 104.53081154823303)
        };
        \addlegendentry{Original}

        \addplot[
            thick,
            orange,
            mark=square*,
            mark size=0.8pt,
        ]
        coordinates {
            (1, 0.09152626991271973) (2, 0.17242884635925293) (3, 0.0913839340209961) (4, 0.0939946174621582) (5, 0.17917203903198242) (6, 0.282102108001709) (7, 0.2017979621887207) (8, 0.2885315418243408) (9, 0.6684606075286865) (10, 0.6347448825836182) (11, 0.4421107769012451) (12, 0.8059351444244385) (13, 0.7406899929046631) (14, 0.5238451957702637) (15, 0.8568167686462402) (16, 0.666942834854126) (17, 1.0771863460540771) (18, 0.7042925357818604) (19, 1.3019890785217285) (20, 1.07706618309021) (21, 0.8976287841796875) (22, 0.873999834060669) (23, 0.6928610801696777) (24, 0.7741765975952148) (25, 1.718909502029419) (26, 1.975466251373291) (27, 1.8110992908477783) (28, 0.5655364990234375) (29, 0.8928515911102295) (30, 1.8343751430511475) (31, 1.7845311164855957) (32, 1.8655719757080078) (33, 1.3073065280914307) (34, 1.966191291809082) (35, 1.4174551963806152) (36, 1.707423210144043) (37, 0.9803948402404785) (38, 2.4625000953674316) (39, 2.158094644546509) (40, 2.881845474243164) (41, 2.905735731124878) (42, 1.459423303604126) (43, 1.388153076171875) (44, 2.123542547225952) (45, 0.5724358558654785) (46, 2.934621572494507) (47, 1.833996057510376) (48, 3.0955209732055664) (49, 1.4978346824645996) (50, 5.249530553817749) (51, 3.071636199951172) (52, 5.071615219116211) (53, 5.596842050552368) (54, 5.3458092212677) (55, 7.147716760635376) (56, 1.841630220413208) (57, 5.235157489776611) (58, 6.010064363479614) (59, 6.803250312805176) (60, 1.7418622970581055) (61, 2.3284175395965576) (62, 5.898124933242798) (63, 2.470304012298584) (64, 6.190903425216675) (65, 3.9602508544921875) (66, 4.6026787757873535) (67, 3.5564708709716797) (68, 11.083359956741333) (69, 4.419821262359619) (70, 6.711677551269531) (71, 9.278450012207031) (72, 1.776261329650879) (73, 4.103154182434082) (74, 8.873135805130005) (75, 9.387413501739502) (76, 11.45934009552002) (77, 9.371326923370361) (78, 10.286589860916138) (79, 10.500911712646484) (80, 13.961575984954834) (81, 5.560499429702759) (82, 8.828439474105835) (83, 6.44760799407959) (84, 8.061930418014526) (85, 10.707072257995605) (86, 9.773473262786865) (87, 11.500428915023804) (88, 22.30401349067688) (89, 7.761149168014526) (90, 24.62308359146118) (91, 9.217711448669434) (92, 9.291869878768921) (93, 22.51078748703003) (94, 25.78345036506653) (95, 27.888298273086548) (96, 14.633507490158081) (97, 17.733864068984985) (98, 35.69531846046448) (99, 14.23778223991394) (100, 55.636488914489746)
        };
        \addlegendentry{CES}

        \end{axis}
    \end{tikzpicture}
    
    \caption{Length comparison in terms of construct units (top) and instance retrieval runtime comparison in seconds (bottom), between original and simplified class expressions using CES, generated in Mutagenesis (left) and Carcinogenesis (right). Results are sorted in ascending order from left to right over the original measurement.}
    \label{fig:results}
\end{figure}
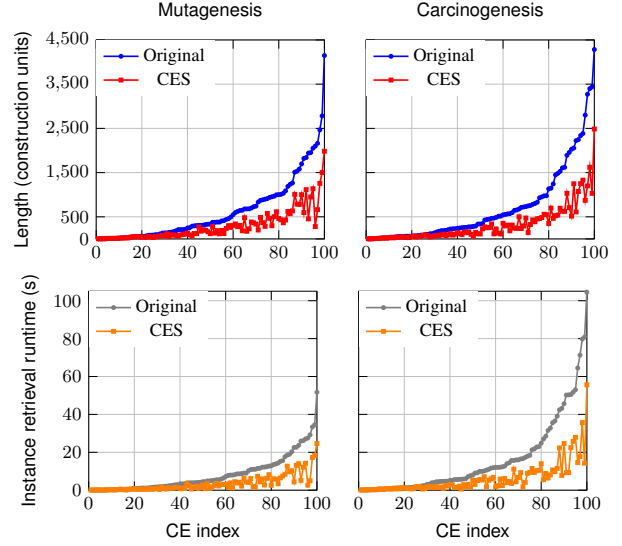

To assess the impact of our syntactic simplification procedure in terms of length reduction and reasoning efficiency, we generated 200 class expressions given random learning problems over two medium-sized datasets, Carcinogenesis and Mutagenesis, from the SML-bench suite~\cite{westphal2019sml}. Complex and lengthy class expressions are generated with the Tree-based \ac{OWL} Class Expression Learner (TDL)~\cite{10.1007/978-3-032-06066-2_29} from the Ontolearn framework~\cite{demir2025ontolearn}, which is known to produce significantly more verbose hypotheses compared to other concept learning algorithms on the framework. 
We refer to the length of a given class expression as the number of the \ac{OWL} constructs required to represent it. 
For reasoning efficiency we measure the runtime of instance retrieval
using StructuralReasoner - the native reasoner
of OWLAPY~\cite{baci2025owlapypythonicframeworkowl}. While each individual rewrite rule used by the algorithm is semantics-preserving, proving correctness for their arbitrary composition would require a global confluence and termination proof, which is outside the scope of this work. Consequently, we evaluate correctness empirically via reasoning-based validation by comparing the sets of instances retrieved for the original and the simplified class expressions and confirming their equivalence.

The results of the experiments are shown in Figure \ref{fig:results}. We recorded the highest length reduction to be 86\% less than the original expression and the highest runtime reduction to be 90\% less than the original. For the conducted experiments the average runtime of simplification is 0.11s on Mutagenesis and 0.16s on Carcinogensis with the longest expression of length 4284 taking 1.34 seconds to simplify.
Our experiments deliberately focused on the TDL, which is known to produce lengthy and complex class expressions. In this setting, the value of our simplifier is particularly evident. For other learners that may generate more compact hypotheses, the gain in simplification may be less pronounced. Nevertheless, the algorithm consistently provides a non-intrusive “bonus” improvement in complexity reduction and reasoning efficiency, since it does not affect the semantics of the resulting expressions.

\section{Discussion and Conclusion}

In this work, we introduced CES, a syntactic simplification algorithm for \ac{OWL} class expressions, and showed its effectiveness in reducing verbosity and improving reasoning efficiency. We highlight the fact that the method can be seamlessly integrated into class expression learning systems (e.g. with TDL), serving as a default post-processing step before presenting the final hypothesis. Since the simplifications are purely syntactic and semantics-preserving, the procedure introduces no risk of information loss, while consistently offering potential benefits with minimal time overhead. The simplification time cost becomes even less significant when considering the fact that class expressions can be reused.

We also acknowledge certain limitations. First, all tested hypotheses were generated by TDL. Consequently, the reported reduction rates may not be representative of concepts produced by other concept learning systems, particularly those that generate shorter or structurally different expressions. 
Another limitation of CES is the order of applying specific processes (e.g., factorization, merging of cardinality restrictions). Since the simplification process proceeds in a single pass, the ordering can influence the outcome. Determining an optimal application strategy remains an open question, and exploring multi-pass or rule-prioritization schemes may yield further improvements.

Despite these limitations, our results demonstrate that substantial reductions in class expression complexity can be achieved, improving reasoning efficiency and reducing verbosity. As such, CES provides a practical step towards more interpretable and maintainable ontology-based systems. At present, CES operates exclusively at the syntactic level. This design choice avoids the need to rely on a reasoner, making the method broadly applicable and independent of the presence of a background ontology. Looking ahead, an exciting direction for future work is the incorporation of semantic simplifications, which could further reduce complexity and uncover additional equivalences beyond purely syntactic transformations.

\bibliographystyle{unsrt}  
\bibliography{references}

\end{document}